# A Surrogate-Assisted Highly Cooperative Coevolutionary Algorithm for Hyperparameter Optimization in Deep Convolutional Neural Network

An Chen, Zhigang Ren, Muyi Wang, Hui Chen, Haoxi Leng, Shuai Liu

School of Automation Science and Engineering, Xi'an Jiaotong University, Xi'an, China

**Corresponding author name**: An Chen

**Affiliation**: School of Automation Science and Engineering, Xi'an Jiaotong University

**Permanent address**: No.28 Xianning West Road, Xi'an Shaanxi, 710049, P.R. China

**Email address**: chenan123@stu.xjtu.edu.cn

# A Surrogate-Assisted Highly Cooperative Coevolutionary Algorithm for Hyperparameter Optimization in Deep Convolutional Neural Network

An Chen, Zhigang Ren, Muyi Wang, Hui Chen, Haoxi Leng, Shuai Liu

School of Automation Science and Engineering, Xi'an Jiaotong University, Xi'an, China

**Abstract:** Convolutional neural networks (CNNs) have gained remarkable success in recent years. However, their performance highly relies on the architecture hyperparameters, and finding proper hyperparameters for a deep CNN is a challenging optimization problem owing to its high-dimensional and computationally expensive characteristics. Given these difficulties, this study proposes a surrogate-assisted highly cooperative hyperparameter optimization (SHCHO) algorithm for chain-styled CNNs. To narrow the large search space, SHCHO first decomposes the whole CNN into several overlapping sub-CNNs in accordance with the overlapping hyperparameter interaction structure and then cooperatively optimizes these hyperparameter subsets. Two cooperation mechanisms are designed during this process. One coordinates all the sub-CNNs to reproduce the information flow in the whole CNN and achieve macro cooperation among them, and the other tackles the overlapping components by simultaneously considering the involved two sub-CNNs and facilitates micro cooperation between them. As a result, a proper hyperparameter configuration can be effectively located for the whole CNN. Besides, SHCHO also employs the well-performing surrogate technique to assist in the hyperparameter optimization of each sub-CNN, thereby greatly reducing the expensive computational cost. Extensive experimental results on two widely-used image classification datasets indicate that SHCHO can significantly improve the performance of CNNs.

## 1. Introduction

By integrating with convolution operations, the convolutional neural network (CNN), one of the most efficient deep learning models, exhibits superior competency to extract meaningful features from the data with local structures, e.g., images and videos [1]. As such, it shows a promising prospect in solving image processing tasks and attracted much research attention in the past decade [2]. To date, CNNs have achieved a considerable number of successful applications. Medical diagnostic [3], electroencephalography signal analysis [4], and autonomous driving [5] are a few typical examples.

Despite the achieved success, it is known that the performance of a CNN highly relies on the network architecture [6, 7], and constructing a suitable architecture for a specific learning task is nontrivial. Most of the existing efficient CNNs, such as the residual convolutional network (ResNet) [8] and the dense convolutional network (DenseNet) [9], are manually designed by experts and then fine-tuned to adapt to the given task through tedious trial-and-error experiments. This process usually requires extensive domain expertise and can be still inefficient with aimless trials. Directly confronting these issues, recent studies aim to automatically locate an efficient CNN architecture in a predefined search space [10-13]. For the block-box characteristic of this problem, the derivative-free reinforcement learning [6, 11] and evolutionary computation (EC) algorithms [7, 12, 13] are widely used as the search paradigms. The latter is increasingly the focus due to its strong search ability and high computational efficiency.

Existing CNN architecture search methods can be mainly divided into two categories, i.e., full-automatic ones and semi-automatic ones. Full-automatic methods adopt some basic CNN layers/blocks as the units and simultaneously optimize the architecture topology (e.g., the number of layers/blocks) and the involved architecture hyperparameters (e.g., the number of convolutional kernels in each convolution layer). The neural architecture search (NAS) network [6], the Q-learning-based meta-modeling neural network (MetaQNN) [14], the automatically evolving CNN (AE-CNN) [15], and the CNN with genetic algorithm (CNN-GA) [16] are several efficient CNNs designed full-automatically. In contrast, semi-automatic methods inherit the topologies of efficient hand-crafted CNNs and focus on optimizing the architecture hyperparameters to improve the learning performance on the given task [17-20]. This type of methods can yield twice the result with half the effort and has achieved various successful applications, including but not limited to lung nodule classification [19] and vehicle logo recognition [21].

Nevertheless, the CNN architecture hyperparameter optimization problem is still a hard nut to crack. For one thing, existing efficient CNN topologies are usually very deep and thus involve many hyperparameters [8, 9], resulting in large search spaces. Conventional optimization methods like EC ones will suffer from the curse of dimensionality and lose their efficacy dramatically [22]. For another, the available number of fitness evaluations (FEs) is usually limited since evaluating each candidate hyperparameter configuration solution requires assessing the performance of the corresponding CNN, where the involved training and validation procedures may cost several hours or even days. A well-performing approach to reducing the FE requirement is to train a surrogate model using some evaluated solutions to approximate the fitness landscape of the optimization problem and then conduct optimization with the assistance of the model [23, 24]. However, this technique can be inefficient on a deep CNN topology. The reason lies in that it is difficult for a surrogate model to describe the high-dimensional problem accurately with limited evaluated solutions.

In an attempt to solve high-dimensional expensive optimization problems, some researchers recently developed the surrogate-assisted cooperative coevolution (SACC) framework [25-27]. SACC first decomposes the original problem into some lower-dimensional subproblems so that the large search space can be greatly reduced and a reliable surrogate model can be efficiently trained for each subproblem. Then these subproblems are cooperatively optimized through a context vector to indirectly solve the original problem. Specifically, SACC concatenates the best sub-solution of each subproblem found so far as the context vector and estimates the fitness of a sub-solution by first inserting it into the corresponding positions of the vector and then evaluating the resulting complete solution with the original problem model [25]. Although SACC has successfully solved many high-dimensional expensive optimization problems [26], it cannot be directly applied to optimizing the architecture hyperparameters in CNNs. As the construction of each surrogate model requires some real-evaluated sub-solutions, SACC still has to assess the whole CNN a certain number of times, which will consume excessive computational resources. Moreover, the layers/blocks in CNNs are generally linked, implying that the hyperparameters involved in adjacent units highly interact with each other. As a result, the interdependency among subproblems will be inevitable, on which the context-vector-based subproblem cooperation mechanism will make SACC converge to a Nash equilibrium rather than a real optimum [28].

Given the above considerations, this study proposes a surrogate-assisted highly cooperative hyperparameter optimization (SHCHO) algorithm for chain-styled CNNs. The architecture topologies of many efficient CNNs can be sketched as a chain of CNN layers/blocks [8, 9]. Based on the overlapping interaction structure among the involved hyperparameters, SHCHO first decomposes them into several overlapping hyperparameter subsets by dividing the whole CNN into some overlapping segments and then constructs a sub-CNN with each segment. This strategy can maintain a complete variable interaction structure in each subproblem, and the generated sub-CNNs can be directly used to evaluate the corresponding sub-solutions, thereby avoiding the expensive invocation of the original model. To effectively coevolve these subproblems, SHCHO employs the extracted feature map of an optimized sub-CNN as the input of its next adjacent one. This information flowing is consistent with that in the original

CNN and achieves macro cooperation among subproblems. Moreover, the algorithm optimizes the overlapping hyperparameters with a multi-objective optimization algorithm by taking the performance of each involved sub-CNN as an objective, which can smoothly settle the potential contradictory configuration requirements and facilitates the micro cooperation between the two adjacent sub-CNNs. To investigate the effectiveness of SHCHO, comprehensive experiments have been conducted on two image classification datasets. The results indicate that SHCHO can help the chained-style CNN achieve competitive classification accuracy compared with state-of-the-art CNN architecture search methods.

In summary, the main contributions of this study are as follows: 1) A surrogate-assisted highly cooperative hyperparameter optimization framework is proposed, which provides a promising solver for fine-tuning the hyperparameters in chain-styled CNNs. 2) A segment-based overlapping decomposition strategy is developed, which tallies with the variable interaction structure in the problem and also provides an efficient evaluation model for each subproblem. 3) A highly cooperative scheme is designed for subproblems, which coevolves subproblems at both macro and micro levels and facilitates the resolution of the original problem. 4) The performance of SHCHO is comprehensively investigated on two image classification datasets, and a thorough analysis of the empirical results is presented.

The remainder of this paper proceeds as follows: Section 2 introduces the background and related works of this study. Section 3 describes the proposed SHCHO algorithm in detail. Section 4 reports experimental settings and results. Finally, Section 5 concludes this paper and discusses some future research directions.

## 2. Preliminaries

### 2.1 *Convolutional Neural Network*

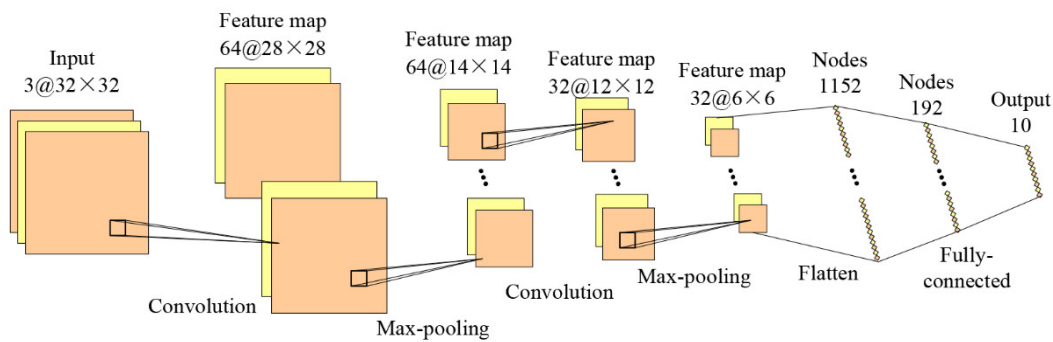

Fig 1. An example of CNN for the CIFAR10 image classification problem.

The CNN model is generally a forward neural network that is apt at tackling learning tasks about images [1]. Fig. 1 presents a CNN example on the CIFAR10 image classification problem. It includes two convolutional layers and two max-pooling layers with each layer further mining the knowledge from the features maps extracted by its previous one. In sequence, the 32×32 input image with three channels is addressed by the convolutional layer with 64 3×3 kernels, the max-pooling layer, the convolutional layer with 32 3×3 kernels, and the max-pooling layer, resulting in a 6×6 feature map with 32 channels. Each pixel in the finally exacted map can be regarded as a valuable feature value of the input image. These pixels are then flattened to an 1152-dimensional vector and input to the two-layer fully-connected network to accomplish the classification task.

The convolutional layer is the key hallmark of CNNs, which consists of lots of convolutional kernels used to extract features from the input image or feature map. As shown in Fig. 2(a), the convolutional kernel traverses the input data and conducts a convolution operation with each covered local region, i.e., bitwise multiplying the kernel and the local region and summing up the results. The convolution results will then be non-linearized with a nonlinear activation function and taken as the output feature maps. In contrast, the pooling layer aims to downsize each feature map while retaining its main features, thus reducing the involved parameters and improving training efficiency. There are two general types of pooling layers, i.e., max-pooling and average-pooling, and a max-pooling example is provided in Fig. 2(b).

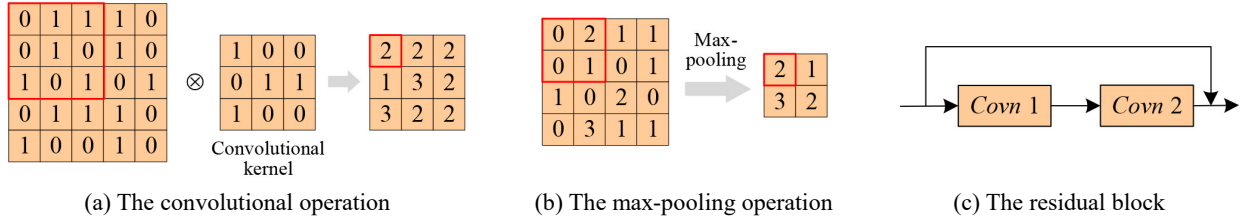

(a) The convolutional operation    (b) The max-pooling operation    (c) The residual block

Fig 2. Examples of some CNN operations and blocks.

Based on the convolutional and pooling layers, some researchers ingeniously designed more complex blocks and developed efficient CNNs accordingly. One of the most well-known blocks is the residual one in ResNet [8]. As shown in Fig. 2(c), it connects the input to the output with two paths, one is traditionally through the convolutional layer, and the other is a shortcut. Consequently, the vanishing gradient problem encountered by conventional CNNs can be effectively avoided. Due to the superiority of the residual block, many variants have been developed, e.g., dense block [9] and inverted residual block [29]. Considering the high efficiency and profound influence, this study adopts ResNet as the optimization object and investigates whether the proposed algorithm can find the architecture hyperparameters that enable the model to achieve better performance than the original one.

## 2.2 CNN Architecture Search Algorithms

The learning ability of a CNN highly depends on its network architecture. How to automatically search for a suitable architecture for a given task remains a research hotspot in recent years. Given the training data $D_{train}$ and the validation data $D_{val}$, the CNN architecture search problem can be formulated as follows:

$$\max_{a \in \mathbb{R}^a, h \in \mathbb{R}^h} f(a, h, w; D_{val}) \\ \text{s.t.} \quad w = \arg\max_{w \in \mathbb{R}^w} f(a, h, w; D_{train}), \quad (1)$$

where $a$, $h$, and $w$ are the architecture topology, the architecture hyperparameter set, and the learning parameter set of the model, respectively, $\mathbb{R}^a$, $\mathbb{R}^h$, and $\mathbb{R}^w$ are the corresponding search spaces, and $f(\cdot)$ measures the model performance, e.g., the classification accuracy for an image classification task. Such an optimization problem is very challenging owing to its black-box and computationally expensive properties.

Some early studies employed reinforcement learning to solve Eq. (1). Zoph et al. [6] exploited a recurrent neural network as the controller to generate CNNs and trained the controller with reinforcement learning to maximize the expected performance of the generated network. The resulting NAS model can achieve competitive performance compared with some efficient hand-crafted architectures. In contrast, Baker et al. [14] constructed a Q-learning agent to pick CNN layers in a predefined search space. The agent first learns from random exploration and then tries to select promising models according to the findings with the $\varepsilon$-greedy strategy. A well-performing model named MateQNN is developed as a result.

Despite the achieved success, reinforcement-learning-based algorithms often require excessive computational resources for the slow convergence speed. By imitating biological evolution in the natural environment, the population-based EC algorithms exhibit impressive global searching ability and have shown great potential for CNN architecture search. As one of the most well-known

EC algorithms, the genetic algorithm (GA) has been widely adopted and studied. For example, Sun et al. [15] constructed an AE-CNN model based on the residual and dense blocks and adopted GA to determine the final network architecture. Afterwards, they further developed a CNN-GA model [16] by introducing skip connections into the design of CNNs. Besides, an asynchronous computational component is also proposed to fully utilize the available computational resources, thereby accelerating the fitness evaluation of individuals.

Due to the fast convergence speed, particle swarm optimization (PSO) has also received much research attention. Wang et al. [30] introduced PSO to CNN architecture search for the first time. They adopted the convolutional and pooling layers as the basic units and encoded the CNN architecture into an internet protocol address format. Huang et al. [31] improved this method with a novel velocity and position updating approach and developed a flexible PSO-based CNN (FPSO-CNN) as a result. More recently, they further developed an efficient PSO algorithm for compact neural architecture search (EPCNAS) [7]. This method adopts the efficient mobile inverted residual block as the building unit and designs a flexible two-level PSO algorithm to evolve both the connections among units and the inner hyperparameter configurations.

In addition to GA and PSO, some other powerful EC algorithms are also employed in the CNN architecture search field. Suganuma et al. automatically constructed CNNs based on Cartesian genetic programming (CGP-CNN) [32]. They adopted a directed acyclic graph with a two-dimensional grid defined on computational nodes to represent a CNN architecture and then evolved the architecture using CGP to maximize the validation accuracy. In contrast, Real et al. [33] maintained a large population, where each individual is encoded as a graph to represent a CNN architecture. Then a set of mutation strategies are applied to evolving the population, resulting in a high-performing CNN called large-scale evolution CNN.

Although the above full-automatic methods have achieved impressive success, it is nontrivial to search for an efficient CNN model from scratch. With tremendous research effort, there are already lots of efficient hand-crafted CNN topologies. For example, the effective ResNet [8] model has gained profound influence and wide applications. It thus can be more effortless to exploit these topologies and optimize the involved architecture hyperparameters to make them adapt to the given learning tasks. This type of optimization problem can be formulated as follows:

$$\begin{aligned} &\max_{\boldsymbol{h} \in \mathbb{R}^h} f(\boldsymbol{h}, \boldsymbol{w}; \boldsymbol{a}, D_{val}) \\ &\text{s.t.} \quad \boldsymbol{w} = \arg\max_{\boldsymbol{w} \in \mathbb{R}^w} f(\boldsymbol{h}, \boldsymbol{w}; \boldsymbol{a}, D_{train}) \end{aligned}. \qquad (2)$$

In fact, there are several classic methods for hyperparameter optimization in the deep learning community [34-38]. Grid search [34] is an enumerative method that tries to test all the configuration combinations and then picks up the best-performing one. Nevertheless, it is inapplicable to Eq. (2) since existing efficient topologies usually involve lots of hyperparameters, resulting in an explosive number of combinations. In contrast, random search [35] randomly samples some candidates from the hyperparameter configuration space and selects the best one. Despite its simplicity, this method has the potential for locating a comparable configuration to grid search with fewer computational resources. Grid search and random search evaluate candidate solutions completely based on the original problem model, which is very expensive for CNN hyperparameter optimization. Bayesian optimization (BO) provides an effective tool for this kind of problems by constructing a surrogate model to approximate the original problem. It first builds a tree-structured parzen estimator (TPE) [36] or Gaussian process (GP) [37] model based on some real-evaluated solutions to fit the original problem and then samples and evaluates the next most promising candidate by optimizing an acquisition function derived from the model. However, the fitting accuracy cannot be guaranteed on a high-dimensional problem.

Some recent studies employed EC algorithms to solve Eq. (2). Wang et al. [18] first enhanced the exploration capability of PSO with a compound normal confidence distribution and then utilized it to explore the CNN hyperparameter configuration space thoroughly. A linear prediction model is also exploited to reduce the expensive FE consumption. Li et al. [20] developed a surrogate-assisted hybrid-model estimation of distribution algorithm (SHEDA) for CNN hyperparameter optimization. To evenly and efficiently search the decision space, SHEDA first initializes a population evenly covering the search space with an orthogonal strategy and then evolves it with a hybrid-model EDA. Besides, the algorithm also employs the GP surrogate model to assist in the optimization process. It has been reported that these methods can significantly improve the performance of some CNNs with several layers. However, they still describe the whole problem with a single surrogate model and thus may lose effectiveness when the model becomes deep.

It can be observed from the above studies that the optimization of CNN architecture hyperparameters is still unfolding, especially for a deep model. The bottleneck lies in how to properly describe the involved large search space with surrogate models. To alleviate this issue, this study draws inspiration from the SACC framework and develops an SHCHO algorithm.

## 3. Proposed method

### 3.1 *Surrogate-Assisted Cooperative Coevolution*

SACC [25, 26] is a newly-developed variant of the influential cooperation coevolution (CC) framework [39]. Taking the divide-and-conquer idea, CC provides a promising solver for the high-dimensional optimization problem by first decomposing it into some lower-dimensional subproblems and then cooperatively optimizing them based on the context vector. However, the context-vector-based sub-solution evaluation method needs to invoke the original high-dimensional problem model frequently, which dramatically limits the application of CC to the problem with high evolution cost. In contrast, SACC incorporates the surrogate technique into the subproblem optimization process and can significantly reduce the real FE requirement.

**Algorithm 1** provides the conventional framework of SACC. After problem decomposition in Line 1, it initializes the context vector **cv** in Line 2 and optimizes the subproblems in Lines 3-11. Specifically, for the current subproblem to be optimized, SACC first samples and evaluates some solutions based on **cv** in Line 5 and then employs them to construct a surrogate model in Line 6. With the assistance of the surrogate model, the current subproblem will be optimized several times in Lines 7-9, where the model will be also continuously refined with the optimization results. Finally, the located best solution will be inserted into the corresponding positions of **cv** in Line 10. The above process iterates until the termination criterion is met, and the final **cv** will be output as the best solution to the original problem.

---

**Algorithm 1**: Conventional SACC

1. Decompose the problem into some low-dimensional subproblems;
2. Initialize the context vector **cv**;
3. **while** the termination criterion is not met
4.     Determine the subproblem *i* to be optimized;
5.     Sample some training solutions and evaluate them based on **cv**;
6.     Train a surrogate model based on these solutions;
7.     **while** the maximum number of optimization times is not reached
8.        Optimize the *i*-th subproblem based on the surrogate model;
9.        Evaluate the optimization results with **cv** and update the surrogate model;
10.    Update **cv** with the located best solution of the *i*-th subproblem;
11.    Re-decompose the original problem if necessary;
12. **return cv**;

The CNN architecture hyperparameter optimization problem is essentially a high-dimensional expensive optimization one and thus can be potentially tackled with SACC. However, there are some gaps to be filled before the application. SACC co-evolves subproblems simply based on the context vector, which requires the interdependencies among subproblems to be minimized so that the theoretical optimum of the original problem can be kept [28]. Nevertheless, this requirement can be hardly fulfilled on the CNN hyperparameter optimization problem since all the hyperparameters are highly linked. Besides, it can be also observed that SACC still requires a considerable number of FEs on the original problem model to construct surrogate models for subproblems, which is unaffordable for Eq. (2). To overcome these difficulties, this study develops the SHCHO algorithm for chain-styled CNNs. Its decomposition process tally with the variable interdependency structure of the optimization problem and can also provide an efficient evaluation model for each subproblem. These subproblems are then highly cooperatively optimized to achieve the resolution of the original problem. We will introduce them sequentially in the next.

3.2 *Decomposition process*

The diagrammatic sketch of a chain-styled CNN $C_{\text{chain}}$ is shown in Fig. 3, where $CLB_i$ ($i = 1,\ldots,n$) denotes the $i$-th CNN layer/block, and $\boldsymbol{h}_i$ is the involved architecture hyperparameter set. Notably, fully-connected layers are not considered in this study since they can easily cause overfitting due to the full-connection nature and also introduce extra hyperparameters. The widely-used softmax classifier [40] will be employed to play the classification role. From Fig. 3, we can observe that the layers/blocks in $C_{\text{train}}$ are linked one by one. As the hyperparameters in each layer/block naturally interact with each other, the whole hyperparameter optimization problem is thus inherently nonseparable, posing a big challenge to CC or SACC. Most existing methods apply either random decomposition [41] or interdependency-learning-based grouping algorithms [42] to this type of problems. However, the former ignores variable interdependency, and the latter will assign all the variables to the same group and thus cannot reduce the problem dimensionality.

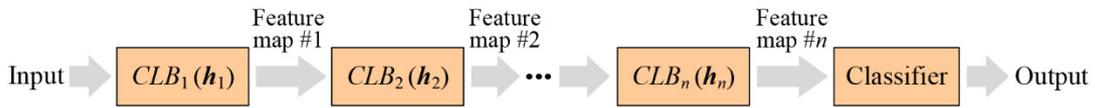

Fig. 3. The diagrammatic sketch of a chain-styled CNN $C_{\text{chain}}$.

Fortunately, the CNN hyperparameter optimization problem has a special interaction structure, which provides an opportunity to decompose it while retaining variable interdependency. A closer inspection of Fig. 3 can find that the hyperparameters from two adjacent layers/blocks directly interact with each other, but those from nonadjacent ones are only indirectly nonseparable (i.e., directly separable). For example, the hyperparameters in $\boldsymbol{h}_{i-1}$ are directly nonseparable from those in $\boldsymbol{h}_i$ and indirectly nonseparable from those in $\boldsymbol{h}_{i+1}$ due to the linkage of $\boldsymbol{h}_i$. Therefore, the optimization problem has an overlapping interaction structure, which means that the hyperparameters can be regarded as coming from multiple subcomponents with shared variables. Given this situation, this study develops a segment-based overlapping decomposition (SOD) strategy. As its name implies, SOD divides $C_{\text{chain}}$ into multiple overlapping segments $SC_1,\ldots, SC_m$ and takes the hyperparameter set in each segment as a subproblem. This decomposition maintains the interdependency integrity among variables in each subproblem and also significantly reduces its dimensionality so that these subproblems can be optimized more effectively.

On the other hand, the real evaluation of a sub-solution in conventional SACC methods needs to train and validate the whole CNN and thus is very expensive. Directly against this limitation, SOD further adds a classifier to each segment and employs the resulting sub-CNN to evaluate its candidate sub-solutions. For each sub-solution, its fitness can be estimated by first updating the hyperparameters in the corresponding sub-CNN accordingly and then assessing the updated model. This evaluation can be more efficient since each sub-CNN contains much fewer layers/blocks than the whole CNN. It is worth mentioning that the above process will be affected by the training and validation data as expressed by Eq. (2). We will describe how to coordinate them for each sub-CNN to achieve a reliable evaluation in the next subsection.

**Algorithm 2** provides the whole pseudocode of SOD. The input parameter $\varepsilon$ controls the number of hyperparameters in each subproblem and will be specified according to the performance of the employed optimizer. The outputs *hgroups* and *scgroups* store the obtained hyperparameter groups and the corresponding sub-CNNs, respectively. After performing initialization in Line 1, SOD decomposes the problem in Lines 2-10, where an indicator *ind* is set to assist in this process. For the current segment *SC* and the hyperparameter group $\boldsymbol{h}_{sc}$, SOD continuously adds $CBL_{ind}$ and $\boldsymbol{h}_{ind}$ in $C_{chain}$ to them, respectively, with *ind* increasing one by one in Lines 3-7 until the number of hyperparameters in $\boldsymbol{h}_{sc}$ exceeds $\varepsilon$. The final *SC* and $\boldsymbol{h}_{sc}$ will be stored in *scgroups* and *hgroups* in Line 8. Next, *ind* will be subtracted by 20% of the number of layers/blocks in *SC* in Line 10 to make the current subproblem and its next one overlapping. Our preliminary experiments show that such an overlapping degree is conducive to the cooperation of the two subproblems. After all the subproblems are generated, a softmax classifier [40] is finally added to each segment in *scgroup* to provide an evaluation model for the corresponding subproblem in Line 11. To better understand the process of SOD, a practical decomposition example is given in Fig. 4.

**Algorithm 2**: $(hgroups, scgroups) \leftarrow \text{SOD}(\varepsilon)$

1. Perform initialization: $hgroups \leftarrow \varnothing$, $scgroups \leftarrow \varnothing$, $ind \leftarrow 1$, and $n \leftarrow$ the number of layers/blocks in $C_{chain}$;
2. **while** $ind \leq n$
3.   Set *SC* and $\boldsymbol{h}_{sc}$ to $CLB_{ind}$ and $\boldsymbol{h}_{ind}$ in $C_{chain}$, respectively;
4.   Update: $ind \leftarrow ind + 1$;
5.   **while** $|H| < \varepsilon$ **and** $ind \leq n$
6.    Concatenate *SC* and $CLB_{ind}$ and add $\boldsymbol{h}_{ind}$ to $\boldsymbol{h}_{sc}$;
7.    Update: $ind \leftarrow ind + 1$;
8.   Update: $hgrups \leftarrow hgrups \cup \boldsymbol{h}_{sc}$ and $scgrups \leftarrow scgrups \cup SC$;
9.   **if** $ind \leq n$
10.    Subtract 20% of the number of layers/blocks in *SC* from *ind*;
11. Add a softmax classifier to each element in *scgroups*;
12. **return** *hgroups* and *scgroups*.

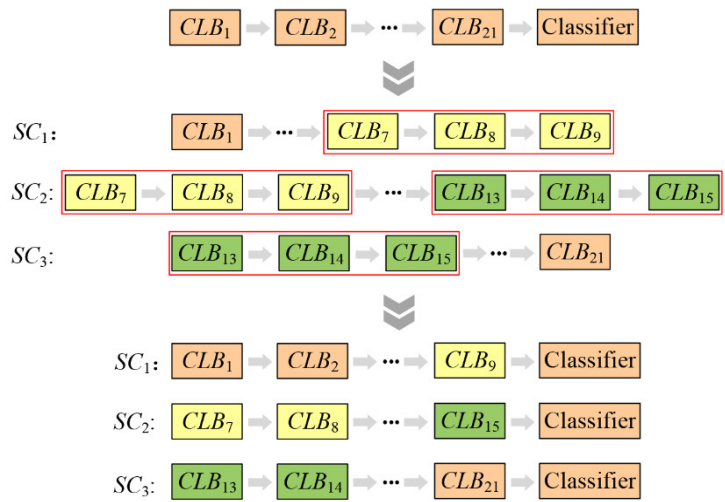

Fig. 4. An example of SCOD.

### 3.3 Subproblem cooperation

After the SOD process, each subproblem involves much fewer architecture hyperparameters, and the involved sub-solutions can be also efficiently evaluated via the corresponding sub-CNN. The next task is to properly co-evolve them to locate a satisfactory hyperparameter configuration for the whole CNN. To this end, SHCHO designs a highly cooperative scheme for the subproblems, which mainly involves two mechanisms, i.e., macro cooperation and micro cooperation.

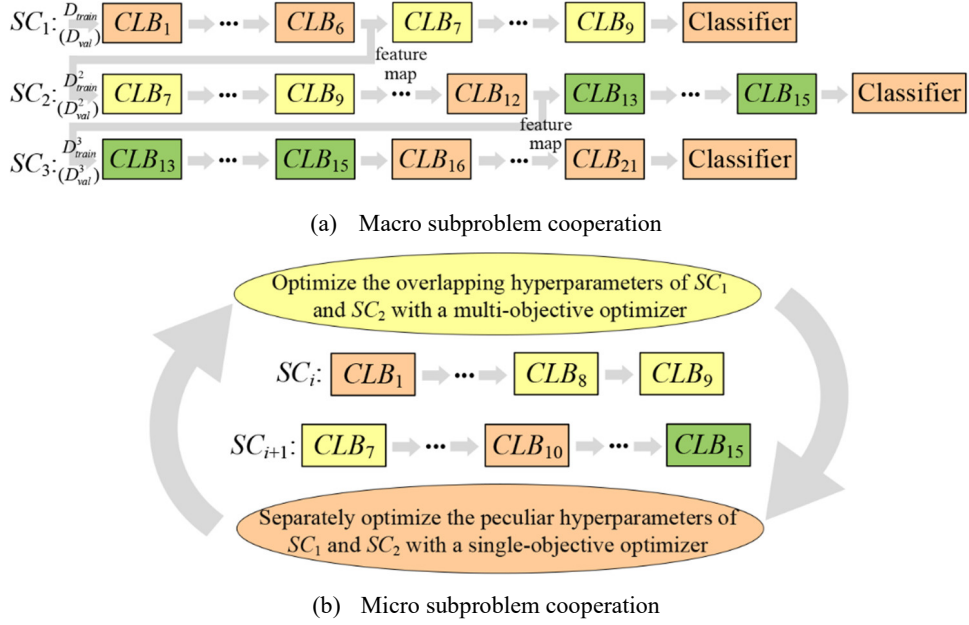

(a) Macro subproblem cooperation

(b) Micro subproblem cooperation

Fig. 5. The highly cooperative scheme for subproblems.

Fig. 5 describes the designed subproblem cooperation scheme, where the generated subproblems in Fig. 4 are taken as examples. As shown in Fig. 5(a), the macro cooperation mechanism aims to coordinate the training data $D_{train}^i$ and the validation data $D_{val}^i$ for each sub-CNN $SC_i$ so that a reliable fitness evaluation of the involved sub-solution can be achieved. As a more straightforward alternative, the given training data $D_{train}$ and validation data $D_{val}$ of the learning task can be directly employed by all the sub-CNNs. However, the final optimization results can only homogeneously improve the performance of these networks on the target task, and the obtained complete hyperparameter configuration can hardly enable the whole CNN to adapt to it. Following the information flow in $C_{train}$, macro cooperation adaptively specifies $D_{train}^i$ and $D_{val}^i$ according to the position of $SC_i$ in $C_{train}$. Concretely, it first employs $D_{train}$ ($D_{val}$) as the training (validation) data of the first sub-CNN $SC_1$, and then adopts the corresponding output feature maps of the last peculiar layer/block in $SC_i$ as $D_{train}^{i+1}$ ($D_{val}^{i+1}$) of $SC_{i+1}$, where the label of each feature map is consistent with its source input in $SC_1$. Such settings enable all the sub-CNNs to extract deep knowledge from the given data in succession. As a result, the finally obtained complete solution can help the whole CNN achieve satisfactory performance.

Nevertheless, the macro cooperation mechanism is still insufficient. The sub-CNNs in the back of $C_{train}$ have no effect on those in the front, which means that the cooperation among subproblems is nonreciprocal. Moreover, it cannot settle the potential contradiction between the configuration requirements of two adjacent CNNs for their overlapping hyperparameters. Directing against these limitations, SHCHO further designs a micro cooperation mechanism for any two adjacent subproblems as shown in Fig. 5(b). For the current two adjacent sub-CNNs $SC_1$ and $SC_2$ and the involved hyperparameter groups $\bm{h}_{SC_1}$ and $\bm{h}_{SC_2}$, this mechanism successively optimizes the overlapping hyperparameter set $\bm{h}_{SC_{(1)(2)}}^{com}$, the peculiar one $\bm{h}_{SC_1}^{pec}$ of $SC_1$, and the peculiar one $\bm{h}_{SC_2}^{pec}$ of $SC_2$ through the following two optimization stages:

1) Construct a multi-objective optimization problem by taking $\bm{h}_{SC_{(1)(2)}}^{com}$ as the decision variables and the performance of $SC_1$ and that of $SC_2$ as the two objectives, i.e.,

$$\max_{\boldsymbol{h}^{com}_{(1)(2)} \in \mathbb{R}^h} \begin{cases} f_{SC_1}(\boldsymbol{h}^{com}_{(1)(2)}, \boldsymbol{w}_{SC_1}; \boldsymbol{h}^{pec}_{SC_1}, \boldsymbol{h}^{r}_{SC_1}, \boldsymbol{a}_{SC_1}, D^1_{val}) \\ f_{SC_2}(\boldsymbol{h}^{com}_{(1)(2)}, \boldsymbol{w}_{SC_2}; \boldsymbol{h}^{pec}_{SC_2}, \boldsymbol{h}^{r}_{SC_2}, \boldsymbol{a}_{SC_2}, D^2_{val}) \end{cases}, \quad (3)$$

$$\text{s.t.} \quad \boldsymbol{w}_{SC_1} = \arg\min_{\boldsymbol{w}_{SC_1} \in \mathbb{R}^w} f(\boldsymbol{h}^{com}_{(1)(2)}, \boldsymbol{w}_{SC_1}; \boldsymbol{h}^{pec}_{SC_1}, \boldsymbol{h}^{r}_{SC_1}, \boldsymbol{a}_{SC_1}, D^1_{val})$$

$$\boldsymbol{w}_{SC_2} = \arg\min_{\boldsymbol{w}_{SC_2} \in \mathbb{R}^w} f(\boldsymbol{h}^{com}_{(1)(2)}, \boldsymbol{w}_{SC_2}; \boldsymbol{h}^{pec}_{SC_2}, \boldsymbol{h}^{r}_{SC_2}, \boldsymbol{a}_{SC_2}, D^2_{val})$$

where $\boldsymbol{h}^r_{SC_1}$ ($\boldsymbol{h}^r_{SC_2}$) includes the other hyperparameters in $SC_1$ ($SC_2$). A multi-objective EC algorithm is employed to solve Eq. (3). The knee point in the located Pareto solution set will be set as the final $\boldsymbol{h}^{com}_{(1)(2)}$, which can achieve the maximum optimization of the two objectives [43].

2) Based on the optimization result of Eq. (3), the peculiar hyperparameters of $SC_1$ and $SC_2$ are then optimized by solving the following two problems:

$$\max_{\boldsymbol{h}^{pec}_{SC_1} \in \mathbb{R}^h} f_{SC_1}(\boldsymbol{h}^{pec}_{SC_1}, \boldsymbol{w}_{SC_1}; \boldsymbol{h}^{com}_{(1)(2)}, \boldsymbol{h}^{r}_{SC_1}, \boldsymbol{a}_{SC_1}, D^1_{val}) \quad (4)$$

$$\text{s.t.} \quad \boldsymbol{w}_{SC_1} = \arg\min_{\boldsymbol{w}_{SC_1} \in \mathbb{R}^w} f(\boldsymbol{h}^{pec}_{SC_1}, \boldsymbol{w}_{SC_1}; \boldsymbol{h}^{com}_{(1)(2)}, \boldsymbol{h}^{r}_{SC_1}, \boldsymbol{a}_{SC_1}, D^1_{train})$$

and

$$\max_{\boldsymbol{h}^{pec}_{SC_2} \in \mathbb{R}^h} f_{SC_2}(\boldsymbol{h}^{pec}_{SC_2}, \boldsymbol{w}_{SC_2}; \boldsymbol{h}^{com}_{(1)(2)}, \boldsymbol{h}^{r}_{SC_2}, \boldsymbol{a}_{SC_2}, D^2_{val}) \quad (5)$$

$$\text{s.t.} \quad \boldsymbol{w}_{SC_2} = \arg\min_{\boldsymbol{w}_{SC_2} \in \mathbb{R}^w} f(\boldsymbol{h}^{pec}_{SC_2}, \boldsymbol{w}_{SC_2}; \boldsymbol{h}^{com}_{(1)(2)}, \boldsymbol{h}^{r}_{SC_2}, \boldsymbol{a}_{SC_2}, D^2_{train})$$

By this means, the performance of the two sub-CNNs can be further enhanced.

The above two optimization stages achieve the competition between $SC_1$ and $SC_2$ and can effectively resolve the potential contradictory configuration requirements of the two sub-CNNs for $\boldsymbol{h}^{com}_{SC_{(1)(2)}}$. Since the number of overlapping hyper-parameters is usually very small, the Pareto solution set of Eq. (4) can be easily found. Besides, it should be also pointed out that considering the interdependencies of $\boldsymbol{h}^{com}_{SC_{(1)(2)}}$ with $\boldsymbol{h}^{pec}_{SC_1}$ and $\boldsymbol{h}^{pec}_{SC_2}$, SHCHO will iteratively conduct the above two stages to locate proper solutions for them. The detailed arrangement for the optimization sequence is described in the next subsection.

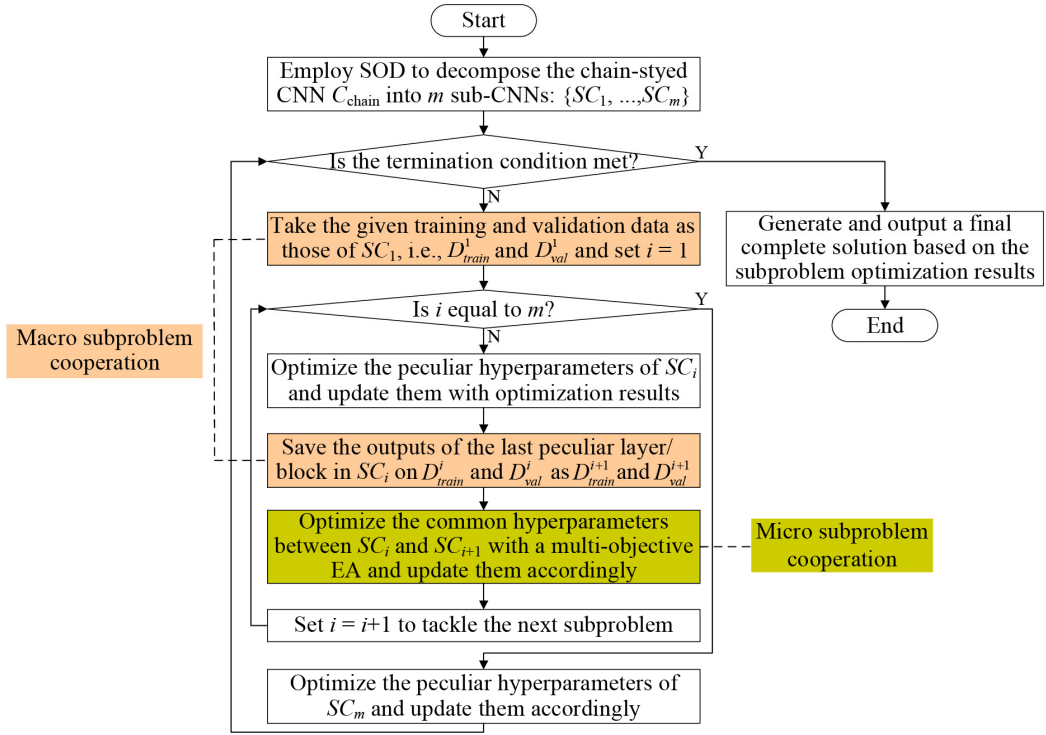

Fig. 6. The overall framework of SHCHO.

### 3.4 Overall Implementation of SHCHO

The overall framework of SHCHO is depicted by Fig. 6. After the SOD procedure, SHCHO then optimizes the hyperparameters in the generated sub-CNNs through the macro and micro coevolution mechanisms. Specifically, it tackles all the subproblems in a round-robin fashion. For the current sub-CNN $SC_i$, the algorithm successively optimizes its peculiar hyperparameters and those shared with the next sub-CNN $SC_{i+1}$ through micro coevolution. Such an optimization sequence arrangement considers the high linkage among all the hyperparameters and is conducive to improve the overall performance of the target chain-styled CNN. Besides, the macro coevolution is incorporated to coordinate the training and validation data of all the sub-CNNs. The above process continues until the termination condition is met, and the complete solution concatenated by the optimization results of all the subproblems will be employed to guide the hyperparameter configuration of the whole CNN.

---

**Algorithm 3**: Surrogate-assisted Highly Cooperative Hyperparameter Optimization

1. Decompose $C_{\text{chain}}$ into *scgroups*: $\{SC_1,\ldots,SC_m\}$ and *hgroups*: $\{\boldsymbol{h}_{SC_1},\ldots,\boldsymbol{h}_{SC_{\_}}\}$ with **Algorithm 2**;
2. Initialize a complete solution $\boldsymbol{h}^*$ to record the optimization results;
3. Randomly spilt the given training data into $D_{\text{train}}$ and $D_{\text{val}}$;
4. **while** the maximum number of iterations is not reached
5.     Set: $D_{\text{train}}^1 \leftarrow D_{\text{train}}$, $D_{\text{val}}^1 \leftarrow D_{\text{val}}$, and $i \leftarrow 1$;
6.     **for** $i = 1 : m-1$
7.       Randomly sample $|\boldsymbol{h}_{SC_i}^{per}|$ solutions for the peculiar hyperparameters $\boldsymbol{h}_{SC_i}^{per}$ in $SC_i$;
8.       Evaluate the solutions based on $SC_i$, $D_{\text{train}}^i$, and $D_{\text{val}}^i$ with **Function 1**;
9.       Employ the localized data generation method to produce more training solutions;
10.      Train an RBF model based on these samples;
11.      **while** the maximum number of optimization times is not reached
12.        Employ a single-objective EC algorithm to locate the optimum of the RBF model;
13.        Evaluate the optimum with **Function 1** and update the RBF model accordingly;
14.      Update the corresponding hyperparameters in $SC_i$ and $\boldsymbol{h}^*$ with the optimization result;
15.      Store the output of the last peculiar layer/block in $SC_i$ on $D_{\text{train}}^i$ ($D_{\text{val}}^i$) as $D_{\text{train}}^{i+1}$ ($D_{\text{val}}^{i+1}$);
16.      Randomly sample $|\boldsymbol{h}_{(i)(i+1)}^{com}|$ solutions for the overlapping hyperparameters $\boldsymbol{h}_{(i)(i+1)}^{com}$ of $SC_i$ and $SC_{i+1}$;
17.      Evaluate the solutions based on $SC_i$, $D_{\text{train}}^i$, and $D_{\text{val}}^i$ with **Function 1**;
18.      Re-evaluate the solutions based on $SC_{i+1}$, $D_{\text{train}}^{i+1}$, and $D_{\text{val}}^{i+1}$ with **Function 1**;
19.      Employ the localized data generation method to produce more training solutions;
20.      Train two RBF models on the two objectives based on these samples, respectively;
21.      **while** the maximum number of optimization times is not reached
22.        Employ a multi-objective EC algorithm to locate the Pareto solution set of the two RBF models;
23.        Evaluate the knee solution on the two objectives with **Function 1** and update the two models accordingly;
24.      Update the corresponding hyperparameters in $SC_i$ and $SC_{i+1}$ and $\boldsymbol{h}^*$ with the knee solution;
25.     Repeat Lines 7-14 to optimize the peculiar hyperparameters $\boldsymbol{h}_{SC_m}^{per}$ and update $\boldsymbol{h}^*$;
26. **return** $\boldsymbol{h}^*$.

---

**Function 1**: *fitness* $\leftarrow$ FitEvalation($sol, SC_i, D_{\text{train}}^i, D_{\text{val}}^i$)

1. Update the hyperparameters in $SC_i$ based on $sol$ and set $fitness \leftarrow 0$;
2. **for** $epoch = 1 : T$
3.    Train $SC_i$ on $D_{\text{train}}^i$ with the SGD optimizer;
4.    Calculate the classification accuracy *temp* of $SC_i$ on $D_{\text{val}}^i$
5.    **if** *temp* > *fitness*
6.      Update: $fitness \leftarrow temp$
7. **return** *fitness*

---

Based on Fig. 6, **Algorithm 3** further provides the detailed pseudocode of SHCHO. After the decomposition and initialization processes in Lines 1 and 2, Line 3 generates the source training and validation data and Lines 4-25 conducts the main optimization procedure. Specifically, for the current sub-CNN $SC_i$, Lines 7-15 first optimize its peculiar hyperparameters through a surrogate-assisted single-objective EC algorithm, where the radial basis function (RBF) in [44] is taken as the basic surrogate model. The number of real-evaluated solutions is set to that of the optimized hyperparameters in Line 7, and the localized data generation method [45] is employed to increase the training solutions in Line 9. This method generates synthetic data within the localized neighborhood of the evaluated individuals and can facilitate building more accurate surrogate models. Our preliminary experiments show that such a configuration can provide enough solutions for training an accurate surrogate model for the low-dimensional subproblem. After the optimization and updating processes in Lines 11-14, Line 15 conducts macro coevolution to provide the training and validation data for the next sub-CNN $SC_{i+1}$. Lines 16-24 are about the micro cooperation process between $SC_i$ and $SC_{i+1}$. Two surrogate models are separately trained on each involved objective sub-CNN in Lines 16-20, and a multi-objective EC algorithm is employed to solve the two-objective optimization problem in Lines 11-13. The above process iterates until the maximum number of iterations is reached, which is set to five in this study. The complete solution that records all the subproblem optimization results is finally output to guide the hyperparameter configuration of the whole CNN.

**Function 1** shows the detailed sub-solution evaluation process. For each sub-solution *sol* to be evaluated on $SC_i$, Line 1 first updates the corresponding hyperparameters in $SC_i$ accordingly. Then $SC_i$ is trained on the training data $D_{train}^i$ for $T$ epochs in Lines 2-6. The best classification accuracy achieved by $SC_i$ on the validation data $D_{val}^i$ during this training process is taken as the fitness value of *sol*. Notably, this study will set $T$ to five as suggested by [20]. It has been indicated that training CNN with several epochs is enough to figure out which CNN is better. Such a configuration can further accelerate the evaluation of each sub-solution.

## 4. Experimental studies

The purpose of this section is two-fold: 1) to evaluate the overall performance of SHCHO; 2) to investigate the effectiveness of the algorithmic components in SHCHO.

### 4.1 *Experimental settings*

In this experiment, we adopted the ResNet [8] with depth=32, denoted as ResNet-32, as the target chain-styled CNN and employed SHCHO to optimize the involved architecture hyperparameters. ResNet-32 is a classic and effective network and contains one conventional layer, fifteen residual blocks, two pooling layers, and one fully-connected layer. Ignoring the less important architecture hyperparameters, we mainly focused on optimizing up to 95 ones, including the pooling type and the kernel size, the number of kernels, and the activation function in each convolutional layer, which result in a large search space. Therefore, the performance of SHCHO can be thoroughly investigated. The available choices and encodings of these optimized hyperparameters are presented in **Table 1**.

Table 1. Settings of the optimized hyperparameters in SHCHO for ResNet-32

| Hyperparameter | Available choices | Encoding |
|---|---|---|
| Number of kernels in the convolutional layer | $\{16,17,\ldots,256\}$ | $\{16,17,\ldots,256\}$ |
| Kernel size in the convolutional layer | $\{3\times3, 5\times5, 7\times7, 9\times9\}$ | $\{0,1,2,3\}$ |
| Kind of activation function in the convolutional layer | {Relu, Sigmoid, Tanh} | $\{0,1,2\}$ |
| Kind of pooling layer | {Max-pooling, Average-pooling} | $\{0,1\}$ |

Two widely-used image classification benchmark datasets, i.e., CIFAR10 [46] and CIFAR100 [46], were used during the experiment. These two datasets both contain $5\times10^4$ training images and $1\times10^4$ testing ones with each image being $32\times32$ RGB format but have different levels of difficulties. Concretely, CIFAR10 consists of ten classes of images, while CIFAR100 covers up to 100 categories. In these two datasets, even the images in the same class have different features like the object shape and location, which poses a big challenge to the classification model. To assess each classification model on them, three popular indicators were employed, i.e., the classification accuracy on the testing dataset, the number of learning parameters in the model, and the GPU days. The third one is equal to the number of used GPUs multiplied by the running days. For example, six GPU days can mean that the algorithm requires two days to terminate with three GPUs. As a result, these three indicators can faithfully measure the model performance, the model size, and the time cost for training the model, respectively, and helps obtain an in-depth assessment of a classification model. To save the expensive computational cost, five independent runs were conducted as suggested in [15], and the best results are reported finally.

As for the algorithm setting, we adopted the classic GA and non-dominated sorting genetic algorithm II (NSGA-II) [47] as the single-objective optimizer and the multi-objective one in SHCHO, respectively. According to their optimization performance, the subproblem size $\varepsilon$ in SHCHO was set to 30. A larger value can be specified for $\varepsilon$ once more efficient optimizers are adopted. During the training process of each sub-CNN, the pre-dominated backpropagation algorithm was used and the stochastic gradient descent method was employed as the optimizer, where the learning rate and the momentum were set to 0.01 and 0.9, respectively. Like other similar studies, the data augmentation techniques [16], including random cropping and horizontal flip, were also applied to the datasets. Then the training dataset was randomly divided into two mutually exclusive sub-datasets when conducting SHCHO. One covers 80% of the images and was taken as the training dataset, and the other was set to the validation dataset.

The best solution located by SHCHO was finally used to guide the hyperparameter configuration in ResNet-32. The resulting network will be trained for 450 epochs on the whole $5\times10^4$ training images and tested on the $1\times10^4$ testing ones. During this training process, we still adopted the stochastic gradient descent optimizer but incorporated a cosine annealing learning rate scheduler with the initial learning rate of 0.1 and the momentum of 0.9 into it to avoid falling into local optimal. The regularization technique, i.e., the cutout method [15], was also used. All the experiments were implemented based on Python 3.8 and PyTorch 1.11.0 and conducted on the computer with an Intel (R) Xeon(R) Platinum 8 CPU and one NVIDIA RTX3080 GPU.

4.2 Comparison with state-of-the-art algorithms

This experiment set aims to comprehensively investigate the overall performance of SHCHO by comparing it with several state-of-the-art methods. To this end, three types of competitors were adopted, i.e., manually designed CNNs, fully-automatically found ones, and semi-automatically found ones. Specifically, the manually designed CNNs include the classic VGG-Net [48], Highway Network [49], ResNets with 32, 110, and 1202 depths [8], and DenseNet [9]. The fully-automatically found CNNs cover two reinforcement-learning-based ones, i.e., NAS [6] and MetaQNN [14], and six EC-based ones, i.e., CGP-CNN [32], Large-scale Evolution CNN [33], AE-CNN [15], CNN-GA [16], FPSO-CNN [31], and EPCNAS-A [7]. As for the semi-automatically found CNNs, BO-TPE [36], BO-GP [37], and SHEDA [20] were the participants. For a fair comparison, the experimental results of these competitors except BO-TPE and BO-GP were directly taken from the original papers. As for the results of BO-TPE and BO-GP, we also taken ResNet-32 as the basic network and implemented them with the open-source Optuna library. The maximum numbers of real FEs were set to 200, where the early stopping policy was also used for each network training process.

**Table 2** presents the classification accuracy, the number of parameters, and the consumed GPU days of each tested model on the CIFAR10 and CIFAR100 datasets. It can be observed that SHCHO achieves 96.10% and 77.03% classification accuracies on them, respectively, and the consumed GPU days are both smaller than three. Such results are very competitive among all the tested methods. Concretely, the following comparison analyses can be made:

Table 2. Comparison with state-of-the-art methods on CIFAR10 and CIFAR100

| Category | Network | CIFAR10 | CIFAR100 | # Parameters | GPU days |
|---|---|---|---|---|---|
| Manually designed | VGG-Net | 93.34 | 71.95 | 20.04M | - |
| | Highway network | 92.28 | 67.61 | - | - |
| | ResNet-32 | 92.49 | 69.17 | 0.46M | - |
| | ResNet-110 | 93.57 | 74.84 | 1.7M | - |
| | ResNet-1202 | 92.07 | 72.18 | 10.2M | - |
| | DenseNet | 94.76 | 75.58 | 1.0M | - |
| Fully-automatic | NAS | 93.99 | - | 2.5M | 22400 |
| | MetaQNN | 93.08 | 72.86 | - | 100 |
| | CGP-CNN | 94.02 | - | 2.64M | 27 |
| | Large-scale Evolution | 94.60 | - | 5.4M | 2750 |
| | Large-scale Evolution | - | 77.00 | 40.4M | 2750 |
| | FPSO-CNN | 93.72 | - | 0.7M | 1.65 |
| | AE-CNN | 95.7 | - | 2M | 27 |
| | AE-CNN | - | 79.15 | 5.4M | 36 |
| | CNN-GA | 96.78 | - | 2.9M | 35 |
| | CNN-GA | - | 79.47 | 4.1M | 40 |
| | EPCNAS-A | 95.05 | - | 0.12M | 1.17 |
| | EPCNAS-A | - | 74.63 | 0.15M | 1.25 |
| Semi-automatic | BO-TPE | 95.90 | - | 19.40M | 2.8 |
| | BO-TPE | - | 75.07 | 21.81M | 3.7 |
| | BO-GP | 96.12 | - | 19.85M | 2.8 |
| | BO-GP | - | 71.64 | 29.28M | 3.7 |
| | SHEDA | 96.36 | - | 10.88M | 0.58 |
| | SHEDA | - | 78.84 | 18.64M | 0.94 |
| | SHCHO | 96.10 | - | 24.68M | 2.2 |
| | SHCHO | - | 77.03 | 30.30M | 2.8 |

1) SHCHO accomplishes significantly better classification results as compared with the classic manually designed CNNs. The classification accuracies of these competitors range from 92.07% to 94.76% on the CIFAR10 dataset and 67.61% to 75.58% on the CIFAR100 one, which are all smaller than those of SHCHO. In particular, SHCHO achieves about 4% and 7% higher accuracies than its basic network ResNet-32 on CIFAR10 and CIFAR100, respectively. Such improvements can clearly illustrate the effectiveness of SHCHO. A closer observation finds that ResNet-110 can also obtain higher classification accuracies compared with ResNet-32 by stacking more residual blocks, but the improvement rates are smaller than those of SHCHO. This indicates that optimizing the hyperparameters of existing well-performing CNN architectures can obtain better results than deepening them. The overfitting issue makes the performance of ResNet-1202 deteriorate compared with ResNet-32 and ResNet-110.

2) As compared with the fully-automatic methods, SHCHO still achieves better results than most of them. It predominates the two reinforcement-learning-based competitors, i.e., NAS and MetaQNN, especially in terms of the time cost of searching for the model. Due to the efficiency of EC and the surrogate technique, its GPU days consumption is fewer than 1% of those of NAS and MetaQNN. When compared with the six EC-based methods, SHCHO is only defeated by CNN-GA on the CIFAR10 dataset and by one more competitor, i.e., AE-CNN, on CIFAR100. Nevertheless, CNN-GA and AE-CNN are less efficient than SHCHO as they both cost more GPU days. It should be pointed out that there is a shortage with SHCHO that the generated networks have lots of learning parameters, especially compared with the recently-developed lightweight EPCNAS-A method. Motivated by the idea in EPCNAS-A, we will focus on incorporating SHCHO into a lighter CNN and introducing the optimization objective of decreasing the number of learning parameters in our future work.

3) Finally, SHCHO also shows competitive performance among the four semi-automatic methods. In comparison with the two Bayesian optimization methods, SHCHO achieves almost similar classification accuracy with them on the CIFAR10 dataset, but significantly outperforms them on the more challenging CIFAR100 one. This result indicates that SHCHO can apply to more difficult classification problems. Equipped with a more efficient optimizer, SHEDA obtains better classification accuracy than

SHCHO, but the gap is small and acceptable.

In summary, the comparisons with state-of-the-art methods have shown the competitive performance of SHEDA. Its effectiveness mainly profits from two factors. First, the SOD strategy fits with the overlapping interaction structure of the hyperparameter optimization problem in CNN and can dramatically reduce the search space of each subproblem. More importantly, the micro and macro cooperation mechanisms achieve systematic coevolution among these subproblems and facilitate the resolution of the original problem. In the next experiment, we will focus on investigating the effectiveness of the two cooperation mechanisms.

4.3 *Investigation of the effectiveness of subproblem cooperation*

For SHCHO, the macro and micro subproblem cooperation mechanisms play vital roles in enhancing its performance. The former coordinates the training and validation data of all the sub-CNNs and enables them to solve the original problem shoulder to shoulder. In contrast, the latter alleviates the contradictory requirements of two adjacent subproblems on the overlapping hyperparameters and facilitates their information exchange. It is thus necessary to empirically investigate the effectiveness of the two cooperation mechanisms. For this purpose, an ablation experiment set was conducted in this subsection.

Specifically, three SHCHO variants, i.e., SHCHO$_{-macro}$, SHCHO$_{-micro}$, and SHCHO$_{-cop}$ were first generated. As the name implies, the macro subproblem cooperation mechanism is removed in SHCHO$_{-macro}$, and the given training and validation datasets of the target classification task are directly taken as those of each sub-CNN for the algorithm completeness. SHCHO$_{-micro}$ deletes the micro cooperation mechanism. To avoid the potential contradictory configuration requirements, it replaces the overlapping decomposition style with an exclusive one. As for SHCHO$_{-cop}$, it further removes both the two cooperation mechanisms and can facilitate the investigation from another perspective. Then a comparison was conducted among the three variants and SHCHO. It is notable that only CIFAR100 was employed during this experiment to save computational resources. As this dataset is very challenging, the performance difference among the three methods can be remarkably uncovered.

**Table 3. Comparison with the variants of SHCHO on CIFAR100**

| Methods | Accuracy | # Parameters | GPU days |
|---|---|---|---|
| SHCHO | 77.03 | 30.30M | 2.8 |
| SHCHO$_{-macro}$ | 75.76 | 31.92M | 2.7 |
| SHCHO$_{-micro}$ | 74.70 | 29.76M | 2.0 |
| SHCHO$_{-cop}$ | 71.16 | 28.41M | 1.9 |
| ResNet-32 | 69.17 | 0.46M | - |

**Table 3** presents the experimental result of the four methods. To provide a comparison baseline, the classification result of ResNet-32 is also given. It can be observed that SHCHO completely dominates its three variants and ResNet-32, and SHCHO$_{-macro}$ and SHCHO$_{-micro}$ also achieve better results than SHCHO$_{-cop}$. These results bi-directionally demonstrate the effectiveness of the two cooperation mechanisms.

In comparison with SHCHO$_{-macro}$, SHCHO consumes similar GPU days to it but achieves 1.27% higher classification accuracy with almost the same numbers of learning parameters. The performance deterioration of SHCHO$_{-macro}$ can be attributed to that without macro cooperation, the evaluated fitness value of a sub-solution can be inconsistent with that on the whole problem, resulting in an improper search direction. The effectiveness of the macro subproblem cooperation mechanism can thus be verified.

When compared with SHCHO$_{-micro}$, SHCHO still shows superior performance. It obtains 2.33% higher classification accuracy than SHCHO$_{-micro}$ with the similar numbers of model parameters. This performance gap indicates that only with macro cooperation, the subproblems cannot exchange their information sufficiently, which impedes their optimization performance. The micro cooperation mechanism is thus necessary. Besides, the effectiveness of the SOD strategy can be also revealed to some extent since SHCHO$_{\text{-micro}}$ employs an exclusive decomposition one. It should be pointed out that SHCHO$_{\text{-micro}}$ consumes the minimum number of GPU days among the three methods. The reason lies in that it involves no the multi-objective optimization process, which will consume extra computational resources. Nevertheless, this consumption is significant and also acceptable.

On the other hand, SHCHO$_{\text{-macro}}$ and SHCHO$_{\text{-micro}}$ also achieve better results than SHCHO$_{\text{-cop}}$. Without the remaining micro (or macro) subproblem cooperation mechanism, the classification accuracy of SHCHO$_{\text{-macro}}$ (or SHCHO$_{\text{-micro}}$) decreases by 4.6% (or 3.54%). The superiorities of SHCHO$_{\text{-macro}}$ and SHCHO$_{\text{-micro}}$ over SHCHO$_{\text{-cop}}$ can further justify the above inferences. A closer inspection finds SHCHO$_{\text{-cop}}$ still accomplishes better classification accuracies than its basic network Res-32. This result indicates that there is much improvement room with the classical hand-crafted CNNs and the importance of architecture hyperparameter optimization.

## 5. Conclusions

This paper proposes a surrogate-assisted highly cooperative hyperparameter optimization (SHCHO) algorithm for chain-styled CNNs. To solve the high-dimensional and computationally expensive CNN architecture hyperparameter optimization problem, SHCHO follows the efficient surrogate-assisted cooperative coevolution and makes two improvements considering the problem characteristics. Specifically, it first develops a segment-based overlapping decomposition strategy according to the hyperparameter interaction structure in chain-styled CNNs. This strategy can maintain the variable interdependencies in each subproblem and provide a sub-CNN to evaluate its sub-solutions efficiently. Then a highly cooperative scheme involving macro and micro cooperation mechanisms is designed to coevolve the generated subproblems. Macro cooperation enables all the sub-CNNs to solve the target learning task shoulder to shoulder following the information flow in the whole CNN, and micro one can settle the contradictory requirements of two adjacent subproblems on the overlapping hyperparameters and facilitates their information exchange. By this means, the original optimization problem can be effectively solved. Extensive experiments have been conducted on two widely-used image classification datasets. The empirical results indicate the competitive performance of SHCHO as compared with some state-of-the-art methods and the effectiveness of the algorithmic components in SHCHO.

In our future work, we will first improve SHCHO to make it applicable to more general CNNs, not limited to the chain-styled ones. Besides, it is also worth pursuing to apply SHCHO to a compact CNN and introduce the optimization objective of decreasing the number of learning parameters, which can significantly extend the practical applicability of SHCHO.

## Credit authorship contribution statement

**An Chen**: Conceptualization, Methodology, Software, Validation, Investigation, Writing-original draft, Writing-review & editing. **Zhigang Ren**: Conceptualization, Methodology, Visualization, Investigation, Writing-review & editing. **Muyi Wang**: Software, Validation, Visualization, Investigation, Writing-review & editing. **Hui Chen**: Software, Investigation, Writing-review & editing. **Haoxi Leng**: Writing-review & editing. **Shuai Liu**: Visualization, Writing-review & editing.

## Declaration of Competing Interest

The authors declare that they have no known competing financial interests or personal relationships that could have appeared to influence the work reported in this paper.

**Acknowledgements**

This work was supported by the National Natural Science Foundation of China (grant number 61873199), the Natural Science Basic Research Plan in Shaanxi Province of China (grant number 2020JM-059), and the Fundamental Research Funds for the Central Universities [grant numbers xzy022020057]

**References**

[1] Y. LeCun, Y. Bengio, and G. E. Hinton, Deep learning, *Nature*, 521 (2015) 436-444, doi: https://doi.org/10.1038/nature14539.

[2] Z. Li, F. Liu, W. Yang, S. Peng, and J. Zhou, A survey of convolutional neural networks: analysis, applications, and prospects, *IEEE Trans. Neural Netw. Learn. Syst.*, 33 (12) (2022) 6999-7019, doi: https://doi.org/10.1109/TNNLS.2021.3084827.

[3] F. E. Fernandes and G. G. Yen, Automatic searching and pruning of deep neural networks for medical imaging diagnostic, *IEEE Trans. Neural Netw. Learn. Syst.*, 32 (12) (2021) 5664-5674, doi: https://doi.org/10.1109/TNNLS.2020.3027308.

[4] Z. Gao, W. Dang, X. Wang, X. Hong, L. Hou, K. Ma, and M. Perc, Complex networks and deep learning for EEG signal analysis, *Cognit. Neurodyn.*, 15 (2021) 369-388, doi: https://doi.org/10.1007/s11571-020-09626-1.

[5] Y. Chen, F. Liu, and K. Pei, Cross-modal matching CNN for autonomous driving sensor data monitoring, in: Proceedings of the IEEE/CVF International Conference on Computer Vision Workshops, 2021, pp. 3110-3119, doi: https://doi.org/10.1109/ICCVW54120.2021.00346.

[6] B. Zoph and Q. V. Le, Neural architecture search with reinforcement learning, in: Proceedings of the International Conference on Learning Representations, 2017, pp. 1-16.

[7] J. Huang, B. Xue, Y. Sun, M. Zhang, and G. G. Yen, Particle swarm optimization for compact neural architecture search for image classification, *IEEE Trans. Evol. Comput.*, in press, doi: https://doi.org/10.1109/TEVC.2022.3217290.

[8] K. He, X. Zhang, S. Ren, and J. Sun, Deep residual learning for image recognition, in: Proceedings of the IEEE Conference on Computer Vision and Pattern Recognition, 2016, pp. 770-778, doi: https://doi.org/10.1109/CVPR.2016.90.

[9] G. Huang, Z. Liu, L. V. D. Maaten, and K. Q. Weinberger, Densely connected convolutional networks, in: Proceedings of the IEEE Conference on Computer Vision and Pattern Recognition, 2017, 2261-2269, doi: https://doi.org/10.1109/CVPR.2017.243.

[10] E. G. Talbi, Automated design of deep neural networks: A survey and unified taxonomy, *ACM Comput. Surv.*, 54 (2) (2022) 1-37, doi: https://doi.org/10.1145/3439730.

[11] Y. Jaafra, J. L. Laurent, A. Deruyver, and M. S. Naceur, Reinforcement learning for neural architecture search: A review, *Image Vis. Comput.*, 89 (2019) 57-66, doi: https://doi.org/10.1016/j.imavis.2019.06.005.

[12] X. Zhou, A. K. Qin, M. Gong, and K. C. Tan, A survey on evolutionary construction of deep neural networks, *IEEE Trans. Evol. Comput.*, 25 (5) (2021) 894-912, doi: https://doi.org/10.1109/TEVC.2021.3079985.

[13] Z. Zhan, J. Li, and Jun Zhang, Evolutionary deep learning: A survey, *Neurocomputing*, 483 (2022) 42-58, doi: https://doi.org/10.1016/j.neucom.2022.01.099.

[14] B. Baker, O. Gupta, N. Naik, and R. Raskar, Designing neural network architectures using reinforcement learning, in: Proceedings of the International Conference on Learning Representations, 2017, pp. 1-18.

[15] Y. Sun, B. Xue, M. Zhang, and G. G. Yen, Completely automated CNN architecture design based on blocks, *IEEE Trans. Neural Netw. Learn. Syst.*, 31 (4) (2020) 1242-1254, doi: https://doi.org/10.1109/TNNLS.2019.2919608.

[16] Y. Sun, B. Xue, M. Zhang, G. G. Yen, and J. Lv, Automatically designing CNN architectures using the genetic algorithm for image classification, *IEEE Trans. Cybern.*, 50 (9) (2020), 3840-3854, doi: https://doi.org/10.1109/TCYB.2020.2983860.

[17] I. Ilievski, T. Akhtar, J. Feng, and C. Shoemaker, Efficient hyperparameter optimization for deep learning algorithms using deterministic RBF surrogates, in: Proceedings of the AAAI Conference on Artificial Intelligence, 31(1), 2017, doi: https://doi.org/10.1609/aaai.v31i1.10647.

[18] Y. Wang, H. Zhang, and G. Zhang, cPSO-CNN: An efficient PSO-based algorithm for fine-tuning hyper-parameters of convolutional neural networks, *Swarm Evol. Comput.*, 49 (2019), 114-123, doi: https://doi.org/10.1016/j.swevo.2019.06.002.

[19] M. Zhang, H. Li, S. Pan, J. Lyu, S. Ling, and S. Su, Convolutional neural networks-based lung nodule classification: A surrogate-assisted evolutionary algorithm for hyperparameter optimization, *IEEE Trans. Cybern.*, 25 (5) (2021) 869-882, doi: https://doi.org/10.1109/TEVC.2021.3060833.

[20] J. Li, Z. Zhan, J. Xu, S. Kwong, and J. Zhang, Surrogate-assisted hybrid-model estimation of distribution algorithm for mixed-variable hyperparameters optimization in convolutional neural networks, *IEEE Trans. Neural Netw. Learn. Syst.*, in press, doi: https://doi.org/10.1109/TNNLS.2021.3106399.

[21] F. C. Soon, H. Y. Khaw, J. H. Chuah, and J. Kanesan, Hyper-parameters optimisation of deep CNN architecture for vehicle logo recognition, *IET Intell. Transp. Syst.*, 12 (2018) 939-946, doi: https://doi.org/10.1049/iet-its.2018.5127.

[22] M. N. Omidvar, X. Li, X. Yao, A review of population-based metaheuristics for large-scale black-box global optimization—Part I, *IEEE Trans. Evol. Comput.*, 26 (5) (2022) 802-822, doi: https://doi.org/10.1109/TEVC.2021.3130838.

[23] Y. Jin, H. Wang, T. Chugh, D. Guo, K. Miettinen, Data-driven evolutionary optimization: An overview and case studies, *IEEE Trans. Cyber.*, 23 (3) (2019) 442-458, doi: https://doi.org/10.1109/TEVC.2018.2869001.

[24] H. Tong, C. Huang, L. L. Minku, X. Yao, Surrogate models in evolutionary single-objective optimization: A new taxonomy and experimental study, *Inf. Sci.*, 562 (2021) 414-437, doi: https://doi.org/10.1016/j.ins.2021.03.002.

[25] Z. Ren, B. Pang, M. Wang, Z. Feng, Y. Liang, A. Chen, Y. Zhang, Surrogate model assisted cooperative coevolution for large scale optimization, *Appl. Intel.*, 49 (2) (2019) 513-531, doi: https://doi.org/10.1007/s10489-018-1279-y.

[26] Y. Liang, Z. Ren, L. Wang, H. Liu, and W. Du, Surrogate-assisted cooperative signal optimization for large-scale traffic networks, *Knowledge-Based Syst.*, 234 (2021) 107542, doi: https://doi.org/10.1016/j.knosys.2021.107542.

[27] M. Sun, C. Sun, X. Li, G. Zhang, F. Akhtar, Surrogate ensemble assisted large-scale expensive optimization with random grouping, *Inf. Sci.*, 615 (2022) 226-237, doi: https://doi.org/10.1016/j.ins.2022.09.063.

[28] L. Panait, Theoretical convergence guarantees for cooperative coevolutionary algorithms, *Evol. Comput.*, 18 (4) (2010) 581-615, doi: https://doi.org/10.1162/EVCO_a_00004.

[29] M. Sandler, A. Howard, M. Zhu, A. Zhmoginov, and L. Chen, MobileNetV2: Inverted residuals and linear bottlenecks," in: Proceedings of the Conference on Computer Vision and Pattern Recognition, 2018, pp. 4510–4520. doi: https://doi.org/10.1109/CVPR.2018.00474.

[30] B. Wang, Y. Sun, B. Xue, and M. Zhang, Evolving deep convolutional neural networks by variable-length particle swarm optimization for image classification, in: Proceedings of the IEEE Congress on Evolutionary Computation, 2018, pp. 1-8, doi: https://doi.org/10.1109/CEC.2018.8477735.

[31] J. Huang, B. Xue, Y. Sun, and M. Zhang, A flexible variable-length particle swarm optimization approach to convolutional neural network architecture design, in: Proceedings of the IEEE Congress on Evolutionary Computation, 2021, pp. 934-941, doi: https://doi.org/10.1109/CEC45853.2021.9504716.

[32] M. Suganuma, S. Shirakawa, and T. Nagao, A genetic programming approach to designing convolutional neural network architectures, in: Proceedings of the Genetic and Evolutionary Computation Conference, 2017, pp. 497-504, doi: https://doi.org/10.1145/3071178.3071229.

[33] E. Real, S. Moore, A. Selle, S. Saxena, Y. L. Suematsu, J. Tan, Q. V. Le, and A. Kurakin, Large-scale evolution of image classifiers, in: Proceedings of the International Conference on Machine Learning, 2017, pp. 2902-2911.

[34] Y. A. Lecun, L. Bottou, G. B. Orr, and K. R. Müller, Efficient BackProp, in: *Neural Networks: Tricks of the Trade*, Springer, 2012, pp. 9–48, doi: https://doi.org/10.1007/978-3-642-35289-8_3.

[35] J. Bergstra and Y. Bengio, Random search for hyper-parameter optimization, *J. Mach. Learn. Res.*, 13 (1) (2012) 281–305.

[36] J. Bergstra, Y. Bengio, R. Bardenet, and B. Kégl, Algorithms for hyperparameter optimization, in: Proceedings of the International Conference on Neural Information Processing Systems, 2011, pp. 2546–2554.

[37] J. Snoek, H. Larochelle, and R. P. Adams, Practical Bayesian optimization of machine learning algorithms, in: Proceedings of the International Conference on Neural Information Processing Systems, 2012, pp. 2951–2959.

[38] Y. Shi, H. Qi, X. Qi, and X. Mu, An efficient hyper-parameter optimization method for supervised learning, *Appl. Soft. Comput.*, 126 (2022) 109266, doi: https://doi.org/10.1016/j.asoc.2022.109266.

[39] M. A. Potter, K. A. D. Jong, A cooperative coevolutionary approach to function optimization, in: Proceedings of the Third Conference on Parallel Problem Solving from Nature, 1994, pp. 249–257, doi: https://doi.org/10.1007/3-540-58484-6_269.

[40] N. M. Nasrabadi, Pattern recognition and machine learning, *J. Electron. Imag.*, 16 (4) (2007) 049901. doi: https://doi.org/10.1117/1.2819119.

[41] Z. Yang, K. Tang, and X. Yao, Large scale evolutionary optimization using cooperative coevolution, *Inf. Sci.*, 178 (15) (2008) 2985-2999, doi: https://doi.org/10.1016/j.ins.2008.02.017.

[42] A. Chen, Z. Ren, W. Guo, Y. Liang, and Z. Feng, An efficient adaptive differential grouping algorithm for large-scale black-box optimization, *IEEE Trans. Evol. Comput.*, in press, doi: https://doi.org/10.1109/TEVC.2022.3170793.

[43] J. Branke, K. Deb, H. Dierolf, and Matthias Osswald, Finding Knees in multiobjective optimization, in: Proceedings of the Third Conference on Parallel Problem Solving from Nature, 2004, pp. 722-731, doi: https://doi.org/10.1007/978-3-540-30217-9_73.

[44] J. Zhang and A. C. Sanderson, DE-AEC: A differential evolution algorithm based on adaptive evolution control, in: Proceedings of the IEEE Congress on Evolutionary Computation, 2017, pp. 3824-3830, doi: https://doi.org/10.1109/CEC.2007.4424969.

[45] J. Li, Z. Zhan, C. Wang, H. Jin, and J. Zhang, Boosting data-driven evolutionary algorithm with localized data generation, *IEEE Trans. Evol. Comput.*, 25 (5) (2020) 923-937, doi: https://doi.org/10.1109/TEVC.2020.2979740.

[46] A. Krizhevsky and G. Hinton, Learning multiple layers of features from tiny images, 2009. [Online]. Available: http://www.cs.utoronto.ca/~kriz/learning-features-2009-TR.pdf.

[47] D. Kalyanmoy, A. Pratap, S. Agarwal, and T. Meyarivan, A fast and elitist multiobjective genetic algorithm: NSGA-II, *IEEE Trans. Evol. Comput.*, 6 (2) (2002) 182–197, https://doi.org/10.1109/4235.996017.

[48] K. Simonyan and A. Zisserman, Very deep convolutional networks for large-scale image recognition, in: Proceedings of the International Conference on Learning Representations, 2015, pp. 1-14.

[49] R. K. Srivastava, K. Greff, and J. Schmidhuber, Highway networks, in: Proceedings of the International Conference on Learning Representations, 2015, pp. 1–6.